\documentclass[conference]{IEEEtran}
\IEEEoverridecommandlockouts
\usepackage{cite}
\usepackage{amsmath,amssymb,amsfonts}
\usepackage{algorithmic}
\usepackage{graphicx}
\usepackage{textcomp}
\usepackage{xcolor}

\usepackage{algorithm}
\usepackage{multirow}
\usepackage{url}
\usepackage{booktabs} 

\usepackage{listings}
\usepackage{marvosym}
\usepackage{hyperref}
\hypersetup{
    colorlinks=true,
    linkcolor=blue,
    filecolor=blue,      
    urlcolor=blue,
    citecolor=blue,
}

\def\BibTeX{{\rm B\kern-.05em{\sc i\kern-.025em b}\kern-.08em
    T\kern-.1667em\lower.7ex\hbox{E}\kern-.125emX}}
\begin{document}

\title{CogGuard: Cognitive and Operational Profiling for Proactive Warning in Edge Intelligent Services
}

\author{
\thanks{This work is supported in part by the National Natural Science Foundation of China (NSFC) under Grant 62272050 and Grant 62302048; in part by the Guangdong Key Lab of AI and Multi-modal Data Processing, Beijing Normal-Hong Kong Baptist University, Zhuhai under 2023-2024 Grants sponsored by Guangdong Provincial Department of Education; in part by Institute of Artificial Intelligence and Future Networks and Engineering Center of AI and Future Education, Guangdong Provincial Department of Science and Technology, China; in part by Zhuhai Science-Tech Innovation Bureau under Grant No. 2320004002772, and in part by the Interdisciplinary Intelligence Super Computer Center of Beijing Normal University at Zhuhai. \textit{(Corresponding author: Zhiqing Tang and Weijia Jia.)}}



\IEEEauthorblockN{
Zhi Yao\textsuperscript{2,1},
Weihao Chen\textsuperscript{3},
Zhiqing Tang\textsuperscript{1,\Letter},
Hanshuai Cui\textsuperscript{2,1},
Qianli Ma\textsuperscript{1,3},
Weijia Jia\textsuperscript{1,4,\Letter}, 
Wei Zhao\textsuperscript{5}
}

\IEEEauthorblockA{
\textsuperscript{1}Institute of Artificial Intelligence and Future Networks, Beijing Normal University, Zhuhai, China\\
\textsuperscript{2}School of Artificial Intelligence, Beijing Normal University, Beijing, China\\
\textsuperscript{3}Faculty of Arts and Science, Beijing Normal University, Zhuhai, China\\
\textsuperscript{4}Guangdong Key Lab of AI and Multi-Modal Data Processing, Beijing Normal-Hong Kong Baptist University, Zhuhai, China\\
\textsuperscript{5}Shenzhen University of Advanced Technology, Shenzhen, China\\
\{yaozhi, chenweihao, hanshuaicui, qianlima\}@mail.bnu.edu.cn, 
\{zhiqingtang, jiawj\}@bnu.edu.cn, 
zhaowei@suat-sz.edu.cn
}

}

\maketitle

\begin{abstract}
Proactive warning is an important capability for edge intelligent services, where the system predicts whether a subject will successfully complete an incoming task under strict latency and privacy constraints. Such prediction depends on both long-term static attributes and short-term dynamic states derived from historical interaction logs. Recent Large Language Models (LLMs) offer strong long-context reasoning for constructing structured profiles from these logs, but existing solutions face two challenges for edge deployment: (1) profiling methods are typically domain-specific and lack a reusable abstraction across service scenarios, and (2) fine-tuning alignment models on heterogeneous edge clusters incurs high synchronization overhead due to the variance in input sequence lengths. To address these challenges, we propose CogGuard, a proactive-warning framework for edge intelligent services. CogGuard decouples offline LLM-based profile construction from online Small Language Model (SLM)-based score prediction through a shared static-dynamic profile-to-score pipeline, and instantiates it in two representative scenarios: educational performance warning and operational task outcome warning. For efficient profile construction, we design scenario-specific profiling methods with prefix-aligned KV-cache reuse to reduce repeated encoding overhead. For edge-side model alignment, we propose a length-aware distributed fine-tuning strategy with contrastive regularization to mitigate workload imbalance on heterogeneous clusters. Experiments on education and operation datasets show that CogGuard reduces profile construction time by up to 48\% and distributed fine-tuning time by 19\%, while achieving MAEs of 13.4 and 5.9, respectively, on 100-point-scale warning tasks. In the largest educational setting, CogGuard reduces prediction error by 15.4\% compared with the strongest baseline.
\end{abstract}


\section{Introduction}

Artificial Intelligence (AI) is transforming high-stakes service domains such as educational tutoring and industrial operations. In these high-frequency interactive scenarios, Large Language Models (LLMs) are evolving from instruction execution tools to decision-making agents with cognitive analysis capabilities~\cite{echterhoff-etal-2024-cognitive}. To deliver personalized behavioral analysis and pre-warning, LLMs must construct accurate user cognitive profiles from massive, unstructured interaction logs and identify the underlying causes of errors~\cite{10.1145/3757925,10.1145/3637528.3671810}. However, traditional cloud-based AI services cannot meet the data privacy and real-time response requirements of these scenarios.

Edge intelligence addresses these limitations by offloading model computation from the remote cloud to edge servers, providing lower-latency services while better protecting user data~\cite{10835069,10707432}. In this context, proactive warning is becoming an important capability of edge intelligent services, where the system predicts whether a subject will successfully complete an incoming task under strict latency and privacy constraints. Such prediction depends on both long-term static attributes and short-term dynamic states derived from historical interactions.

However, building a proactive warning service at the edge faces two obstacles. First, constructing accurate profiles from massive historical logs is computationally intensive~\cite{srivatsa2025preble}, yet the resulting multi-level profiles can only be fully interpreted by high-parameter models that are unsuitable for edge deployment~\cite{10.1145/3719664, 11359542}. Second, while fine-tuned Small Language Models (SLMs) offer a lightweight alternative for edge inference, distributed fine-tuning on heterogeneous edge clusters suffers from synchronization bottlenecks that existing methods have not adequately addressed~\cite{lau2025adaptive, 8657771}. Cognitive and operational profiling for proactive warning in edge intelligent services is therefore an urgent need, but two challenges remain.

\textit{The first challenge is how to rapidly and reliably construct structured profiles from time-varying subject states for edge proactive warning services.} Proactive warning requires accurately capturing the impact of both external task entities and users' internal cognitive levels on task outcomes~\cite{lv-etal-2024-coggpt}, a task that requires complex and redundant model computations for cognitive analysis. Existing methods map historical features onto knowledge graphs and use the constructed structured profiles during subsequent services~\cite{10.1145/3589334.3645574}. However, traditional graph construction methods extract all entity relationships without discrimination, weakening the representation quality, and subsequent graph clustering also leads to high latency~\cite{zhong2025semragsemanticknowledgeaugmentedrag}. 
Moreover, existing profiling methods are often domain-specific and lack a semantic interface that can be seamlessly integrated into downstream predictive services. This leads to substantial overhead in re-engineering the alignment pipeline for each new scenario, making it difficult to deploy generalizable proactive-warning services on heterogeneous edge clusters.

\textit{The second challenge is how to mitigate the synchronization bottleneck caused by workload heterogeneity in distributed edge fine-tuning.} Fine-tuning SLMs on resource-constrained edge servers is time-consuming, and the input sequence lengths of different historical logs vary widely, leading to computational imbalance across training servers. Existing distributed training approaches reduce cost by adaptively allocating batch sizes based on hardware heterogeneity~\cite{lau2025adaptive}, but these hardware-centric methods schedule workloads solely according to device computing power without accounting for the variance in training sample complexity. In particular, the high variance of user log sequence lengths causes severe straggler effects, where faster nodes idle while waiting for slower ones that process longer sequences. Efficiently coordinating distributed fine-tuning under such input-level heterogeneity on edge clusters remains an open problem.

\begin{figure}[t]
	\centering
	\includegraphics[width=1\linewidth]{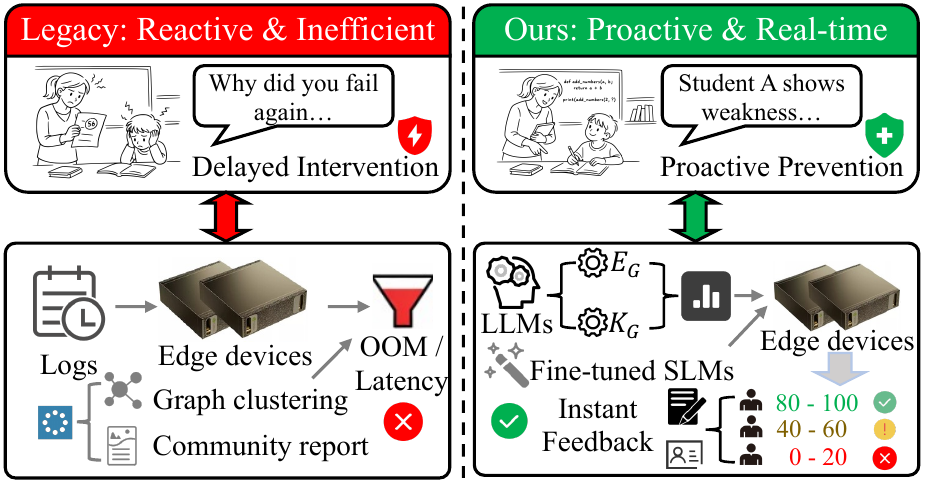}
	\centering
	\caption{Overview of CogGuard.}
	\label{fig:1}
\end{figure}

To address these challenges, we propose CogGuard, as shown in Fig.~\ref{fig:1}, a proactive-warning framework for edge intelligent services that combines structured profiling with efficient distributed model fine-tuning. CogGuard adopts a large-small model collaboration approach to balance reasoning capability and edge inference latency. High-performance LLMs handle the complex offline extraction of structured profiles, while fine-tuned SLMs are deployed on resource-constrained edge nodes for real-time proactive warning. The framework operates in two phases, scenario-specific profiling that constructs structured service contexts from historical logs, and profile-to-score alignment that fine-tunes SLMs with length-aware scheduling and contrastive regularization. Experiments in educational and operational scenarios demonstrate the effectiveness of the proposed pipeline in two representative proactive-warning settings~\cite{10.1145/3777375}.

The central design principle of CogGuard is to decouple expensive scenario-specific profile construction from lightweight shared profile-to-score alignment. The former allows each service scenario to use its own profiling logic, while the latter provides a reusable prediction interface. To make this alignment robust under input and state heterogeneity, CogGuard further introduces structured marker tokens, length-aware training partitioning, and state-sensitivity regularization. Our main contributions are summarized as follows.
\begin{enumerate}

    \item \textbf{Problem Formulation.} We formulate proactive warning in edge intelligent services as a static-dynamic profile-to-score prediction problem, where task outcomes depend jointly on long-term subject attributes, time-varying states, and the current task context.

    \item \textbf{Structured Profiling.} We propose a scenario-specific profiling mechanism. For educational logs, a dual-graph cognitive profiling method decouples user tracking into an entity relationship graph and a knowledge concept graph, with a prefix-aligned KV cache reuse strategy to reduce repeated long-context encoding overhead. For operational logs, static hardware and dynamic runtime profiles are constructed through controlled chaos testing.

    \item \textbf{Efficient Alignment.} We introduce a length-aware distributed fine-tuning strategy that partitions training data according to input sequence lengths to mitigate the straggler problem on heterogeneous edge clusters, and propose a contrastive regularization objective that prevents the model from relying on static problem content rather than personalized profile states.

    \item \textbf{Cross-Scenario Validation.} We instantiate the framework in educational and operational edge service scenarios by constructing specialized datasets tailored to each profiling method. The results show that the shared profile-to-score alignment pipeline can be effectively applied in two distinct proactive-warning settings.

\end{enumerate}

The remainder of this paper is organized as follows. Section~\ref{section_related_work} reviews related work. Section~\ref{section_dataset} details our dataset construction methodology. Section~\ref{section_method} provides a detailed description of our method. The implementation setting and experimental results are described in Section~\ref{section_result}. Finally, Section~\ref{section_conclusion} concludes the paper.

\section{Related Work}
\label{section_related_work}

\subsection{Structured Profiling and Graph RAG}
Traditional knowledge tracing (KT) methods such as DKT~\cite{piech2015deep} achieve high prediction accuracy but lack interpretability. Recent works address this limitation by constructing cognitive graphs through LLMs~\cite{liu2025dualreasoninggnnllmcollaborative,Lv_Liu_Gao_Zhang_Lu_Zhu_2025}. For example, \cite{wu2025embracingimperfectionsimulatingstudents} uses LLM agents to construct cognitive prototypes and simulate students with diverse cognitive levels, but it relies on effective retrieval and cannot directly quantify a student's mastery level. Beyond educational scenarios, Graph RAG is also widely used in operational contexts~\cite{cite-key,wu-etal-2025-medical}. For instance, \cite{MIAO2026114855} applies Graph RAG to fault diagnosis on high-speed trains, and GNN-RAG~\cite{mavromatis-karypis-2025-gnn} achieves lightweight retrieval for multi-hop inference tasks. However, current Graph RAG methods still rely on complex graph clustering and community summary generation, which grow exponentially with the graph size, making them unsuitable for edge applications~\cite{han2025retrievalaugmentedgenerationgraphsgraphrag}.

Moreover, existing approaches do not treat shared entity extraction and personalized state extraction as distinct profiles. Our dual-graph cognitive profiling method addresses this gap by decoupling entity-level and knowledge-level graphs. The resulting profiles are internalized into the fine-tuned model, eliminating the need for graph clustering and retrieval at inference time.

\subsection{Efficient Training on Heterogeneous Clusters}
Fine-tuning SLMs on resource-constrained edge servers is time-consuming. Distributed Data Parallel (DDP) and Ray~\cite{222605} are widely adopted for parallelizing training across multiple nodes, but they require all workers to proceed in lockstep, leading to high synchronization overhead when hardware capabilities differ. To reduce this overhead, HetPipe~\cite{254418} pipelines layers across heterogeneous GPUs to overlap computation between fast and slow devices. Similarly, Accpar~\cite{9065574} partitions computation graphs according to each device's throughput and assigns proportional workloads. Both methods effectively balance hardware heterogeneity, but they schedule work based solely on device computing power.

However, fine-tuning tasks exhibit greater input heterogeneity than pre-training because user historical logs vary widely in sequence length. This variance causes straggler effects even when hardware workloads are balanced, and existing hardware-centric methods do not account for it. Our method addresses this gap by introducing a length-aware data partitioning strategy that groups samples by sequence length before distributing them across servers.

\section{Dataset Curation}
\label{section_dataset}

\subsection{Educational Scenario}

We construct an educational dataset tailored for proactive warning from programming interaction logs. To support structured profile construction in CogGuard, each sample must include three essential elements: a list of the main knowledge concepts involved, sufficient historical records for each student, and detailed problems, solutions, and scores for each record.
Standard knowledge tracing datasets~\cite{NEURIPS2023_8cf04c64,liu2023simplekt} contain only sparse interaction logs, lacking the rich textual context required for LLM-based cognitive reasoning. The prior LLM-generated student datasets~\cite{wu2025embracingimperfectionsimulatingstudents} remain limited in scale.

We select C++ programming as our evaluation scenario~\cite{yang2025codethinkthinkcode}.
The dataset is sourced from the online programming platform NowCoder\footnote{\url{https://ac.nowcoder.com/acm/problem/list/}}.
The resulting dataset contains 10036 submission records covering 1076 distinct problems from up to 40 students. This scale mirrors the typical size of a natural class in real-world educational settings and represents a realistic workload for a single edge-deployed warning node. Each student interacts with multiple problems and may submit multiple attempts for the same problem.
We implement a three-stage quality-control pipeline. (1) \textbf{Data Cleaning.} We filter out excessive repeated submissions for the same issue to avoid model overfitting. (2) \textbf{Standardization.} We retain valid submission materials that reflect dynamic learning states, ensuring each sample includes the student's solution, reference answers, and specific error types. The target scores are derived from the platform and mapped into our prediction intervals. (3) \textbf{Structuring and Split.} The raw data consists of submission records in chronological order. To prevent data leakage, we split the dataset by students rather than randomly by records and feed historical records sequentially to the LLM according to submission timestamps to track the dynamic cognitive evolution accurately.

\begin{table}[t]
\caption{Distribution and Configuration of Fault Injection Scenarios}
\label{tab:fault_injection}
\centering
\begin{tabular}{lllc}
\toprule
\textbf{Category} & \textbf{Configuration} & \textbf{Parameters} & \textbf{Count (\%)} \\
\midrule
\multirow{5}{*}{Network} & packets loss & percent 50 & \multirow{5}{*}{360 (83.5\%)} \\
 & packets corrupt  & percent 50 & \\
 & packets duplicate & percent 50 & \\
 & delay & 3000 ms & \\
 & bandwidth limit & 5 Mbps & \\
\midrule
\multirow{2}{*}{Input File} & file rename & \multirow{2}{*}{NFS file read} & \multirow{2}{*}{288 (66.8\%)} \\
 & replace data in file & & \\
\midrule
\multirow{2}{*}{CPU} & \multirow{2}{*}{stress-cpu} & workers 10 & \multirow{2}{*}{288 (66.8\%)} \\
 & & load 50 & \\
\midrule
\multirow{2}{*}{Disk I/O} & add-payload write & 20G, 8 threads & 216 (50.1\%) \\
 & read payload & 200G, 7 threads & 216 (50.1\%) \\
\midrule
Memory & memory stress & 25 GB & 216 (50.1\%) \\
\bottomrule
\end{tabular}
\end{table}

\subsection{Operational Scenario}
We construct an operations dataset for proactive task outcome warning under runtime disturbances. Although mainstream orchestration platforms like Kubernetes (K8s) support automated deployment and resource management, their default scheduling mechanisms rely on static resource constraints and affinity rules, lacking awareness of runtime server disturbances. As a result, task schedulers based on static configuration cannot cope with unpredictable dynamic failures.

Instead of passively waiting for unpredictable failures, we employ a controlled chaos profiling strategy to actively probe system boundaries and stress-test the servers. This allows the proactive warning system to learn machine behaviors under high pressure before real failures occur. Through this process, we map the hardware and runtime profiles of the target infrastructure. The static hardware profile $P_H$ captures innate attributes such as CPU architecture, GPU capacity, and network bandwidth. The dynamic runtime profile $P_R$ describes the injected fault status at the current time~\cite{10903891}.

\begin{figure*}[t]
  \centering
  \includegraphics[width=0.95\textwidth]{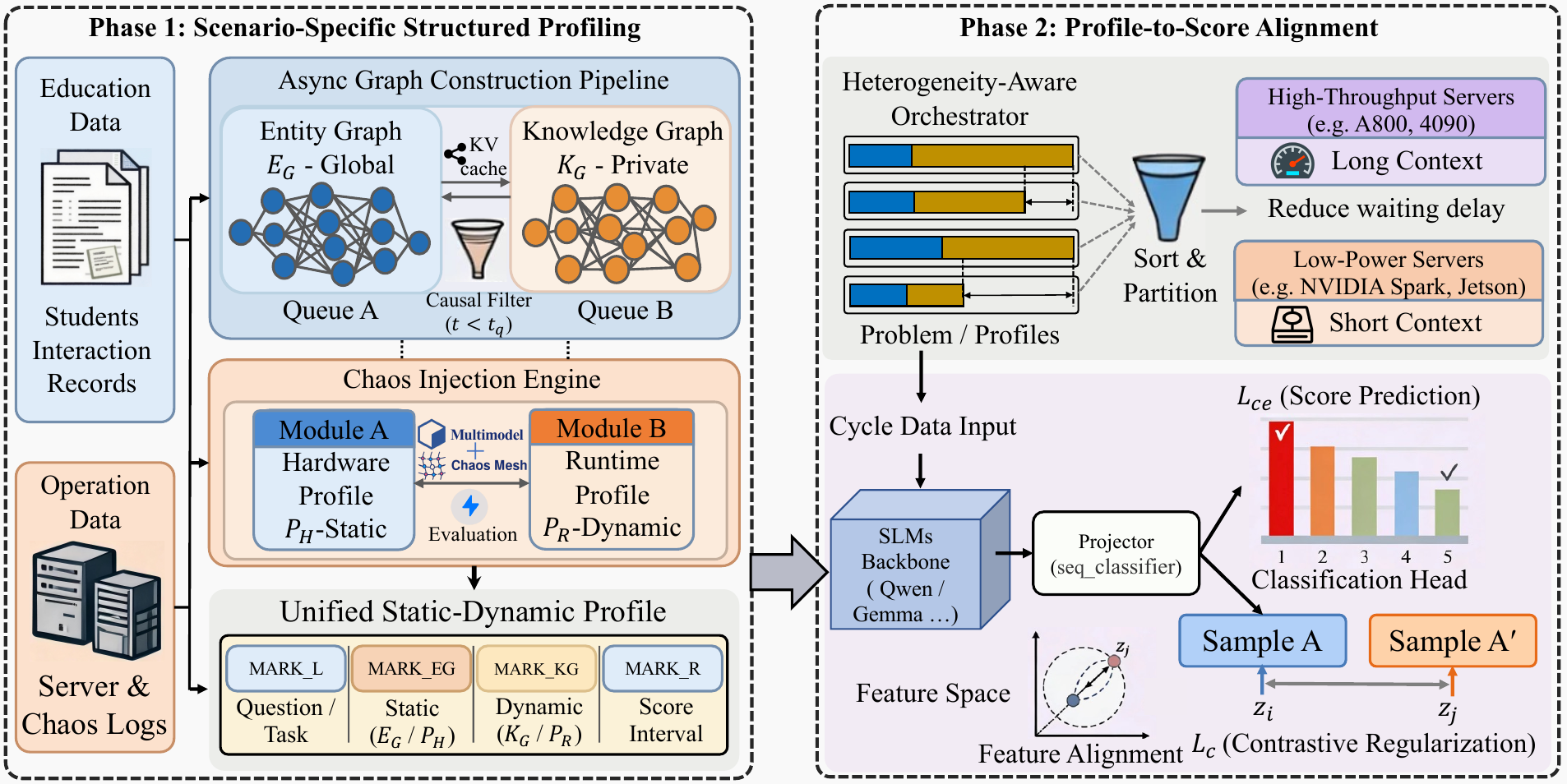}
  \centering
  \caption{Overview of the proposed method. Phase 1 constructs structured profiles via scenario-specific methods: asynchronous dual-graph construction for education and chaos injection for operations. Phase 2 uses a heterogeneity-aware orchestrator for fine-tuning, and knowledge graph summary sensitivity is enhanced through contrastive regularization to optimize score prediction.}
  \label{fig:figure2}
\end{figure*}

We use Chaos Mesh\footnote{\url{https://chaos-mesh.org/}} to inject up to six types of faults and 11 corresponding optional parameters into heterogeneous servers. Table~\ref{tab:fault_injection} lists the specific fault types and their occurrence frequencies. The dataset simulates varying degrees of system instability by injecting multiple concurrent faults into each executed task.
The number of concurrent faults ranges from 1 to 6 and peaks at 4 faults (32.7\%), indicating a moderate concentration around mid-complexity scenarios. Under these conditions, we execute mixed AI inference workloads (including text-to-text and image-to-text tasks) and record the execution results.

\section{Methods}
\label{section_method}

\subsection{Overview}

Before detailing the method, we formally define the proactive warning task. Given a subject $u \in U$ (such as a student or configurable server) and its historical interaction logs $L_u = \{l_1, l_2, \ldots, l_{t-1}\}$, the goal is to predict the performance score $y_t$ of a new incoming task $q_t$ in real time.

CogGuard models proactive warning in different scenarios through a shared profile-to-score prediction abstraction. We learn a lightweight mapping function $f_{\theta}$ parameterized by an edge-deployed SLM: $y_t = f_{\theta}(p_a, p_b, q_t)$, where $p_a$ and $p_b$ denote the structured context extracted from $L_u$. In the educational scenario, they represent shared task entities and personalized knowledge states; in the operational scenario, they represent static hardware profiles and dynamic runtime faults. This design allows the downstream alignment pipeline to be shared across scenarios while keeping the upstream profiling process scenario-specific. 
It is worth noting that CogGuard does not assume a fully unified upstream profiling process. Instead, it provides a reusable downstream alignment interface that converts scenario-specific profiles into a shared static-dynamic profile-to-score prediction format. Therefore, different services may instantiate their own profiling modules, while sharing the same alignment and warning pipeline.

As illustrated in Fig.~\ref{fig:figure2}, the method operates in two phases: (1) Scenario-Specific Structured Profiling, where we employ parallel asynchronous graph construction (for students) or chaos engineering fault injection (for servers) to extract the service context under causal constraints; and (2) Profile-to-Score Alignment, where we fine-tune an SLM to predict scores based on the generated profile summaries. To handle edge heterogeneity and textual bias, we introduce a length-aware workload orchestrator and a contrastive regularization loss.

\subsection{Scenario-Specific Structured Profiling}
We categorize behavioral subjects into two types based on whether their intrinsic characteristics are controllable.

For human-centric subjects such as students, performance is mainly influenced by problem content, familiarity with the problem, and mastery of relevant knowledge concepts. Since the entities in a problem stem may appear in multiple problems, all students share a global entity graph while each student maintains an individual knowledge graph. For configurable servers, task outcomes are mainly affected by task content, hardware configuration, and runtime status. Since servers are machines rather than humans, we can directly model the machine state through chaos experiments and skip graph construction.

\textbf{Human-Centric Subjects.}
To characterize the impact of problem entities and students' intrinsic knowledge on problem-solving outcomes, we propose an asynchronous dual-graph construction method that builds the entity graph $E_{G}$ and student knowledge graph $K_{G}$ in parallel.
Unlike traditional Graph RAG, our method only requires the summarized content generated by graph construction to build student profiles~\cite{NEURIPS2024_efaf1c97,edge2025localglobalgraphrag}.
This decoupling can model two types of prediction: predicting the solutions of the same problem for multiple students' profiles and predicting the solutions of different problems for one profile~\cite{yang2025graphusionragframeworkknowledge,zhang2025leanragknowledgegraphbasedgenerationsemantic}.
By dispatching $E_{G}$ and $K_{G}$ construction to independent queues, we decouple problem semantic analysis from dynamic knowledge assessment, thereby maximizing system throughput.
This dual-graph decoupling also simplifies complex reasoning tasks. Since LLMs often struggle to distinguish between objective problem entities and subjective knowledge concepts within lengthy prompts, separating them improves interpretability and enables the cache prefix optimization described below.

The $E_G$ is a globally shared structure where the interaction data from all students are aggregated into a single unified graph. Since entity concepts are universal, a shared topology allows us to aggregate collective familiarity, where low familiarity indicates high difficulty or novelty across the student population~\cite{SEGAL2019261}. The $K_G$ is a student-specific profile that maintains complete data isolation for each individual.

We design a prefix-aware KV cache reuse mechanism to exploit the shared interaction context across solutions. By standardizing the prompt template so that shared fields appear at the prefix, we enable inference engines to reuse KV caches for the same problem. This reduces the FLOPs required for redundant context encoding while preserving the sequential integrity of learning histories. This cache reuse is applied only during offline profile construction, where repeated LLM encoding of shared problem prefixes dominates the cost. It is orthogonal to the downstream SLM fine-tuning stage, which focuses on profile-to-score alignment.

The shared prefix follows a fixed input order: \texttt{Question: \{question\}}, \texttt{Description: \{desc\}}, \texttt{Student Program: \{program\}}, and \texttt{Error Description: \{error\_desc\}}. For personalized knowledge concept extraction, task-specific instructions (\texttt{Knowledge Points to Extract: \{kp\_list\}}) are appended at the end of this shared prefix.
When processing knowledge graphs, we merge the \texttt{{desc}} field into the \texttt{Question} field. Since knowledge graph construction focuses on evaluating specific knowledge points, the detailed problem description mainly serves as background context. Consolidating it into the question avoids redundant encoding and prevents dilution of the LLM's attention.

To guide the LLM in structured extraction, we design prompt constraints for knowledge graph construction. The model maps the student's program and error descriptions to the predefined C++ knowledge points \texttt{\{kp\_display\}}, as shown in Listing~\ref{lst:kp_display}.
The prompt also requires evaluating the mastery level of each extracted point. By analyzing the compilation or logical error descriptions, the LLM categorizes the student's current state as either demonstrating understanding (``Good'') or lacking comprehension (``Bad''). The code and complete prompt templates are open-sourced in our repository\footnote{\url{https://github.com/Mrzhiyao/CogGuard}}.
\begin{lstlisting}[
    caption={The content of \{kp\_display\} for C++ programs.},
    label={lst:kp_display}
]
Syntax : Input_Output_and_Sequential_Structure Control_Structure

Data_Structure : Linked_List Stack Queue Graph_Structure String_Algorithms

Algorithm : Enumeration_and_Sorting Search Greedy Simulation Binary_Search

Dynamic_Programming Number_Theory Hash Probability_and_Statistics Game_Theory
\end{lstlisting}

Based on these extracted states, the profile summary quantitatively characterizes the student's mastery over entities or knowledge nodes. For any node $m \in E_G \cup K_G$, we first map the LLM's categorical evaluation into a binarized mastery label $y_m \in \{0, 1\}$ (i.e., ``Good'' as 1, ``Bad'' as 0).  We then compute the base mastery rate $R_m = n_{m,1} / (n_{m,1} + n_{m,0})$, where $n_{m,1}$ and $n_{m,0}$ are the historical counts of mastered and unmastered instances.
We also introduce a frequency factor $f_m = n_m / N$ to weight the student's familiarity, where $N$ is the total number of historical records and $n_m$ is the count involving node $m$.
The node score $S(m)$ is calculated by combining the base accuracy and the frequency factor:
\begin{equation}
\label{score}
    S(m) = \begin{cases}
    \mu + f_m \cdot (1 - \mu) & \text{if } R_m \ge \mu \\
    \mu - f_m \cdot \mu & \text{if } R_m < \mu
\end{cases}
\end{equation}
where $\mu$ is the initial neutral score. The system ranks all nodes by $S(m)$ and retains the top-$k$ strongest and weakest nodes to generate the final cognitive profiling summary.

\textbf{Machine-Centric Servers.} While server performance is theoretically deterministic, real-world execution is heavily affected by dynamic disturbances. As detailed in Section~\ref{section_dataset}, we probe system boundaries using controlled chaos injection to construct the static hardware profile $P_H$ and the dynamic runtime profile $P_R$. Rather than employing graph extraction, the model directly takes $P_H$ and $P_R$ as the standardized profile input.

\subsection{Profile-to-Score Alignment}
We align structured profiles to task scores by fine-tuning SLMs. To improve the model's understanding of the input structure and enable prefix sharing during batched inference, we introduce four special tags: $<$MARK\_L$>$, $<$MARK\_EG$>$, $<$MARK\_KG$>$, and $<$MARK\_R$>$.

\textbf{Task Outcome Scoring.} In the educational scenario, the target scores are derived directly from the platform (Section~\ref{section_dataset}). In the operational scenario, we design a scoring mechanism based on mixed AI inference workloads to generate target labels ($y_t$). As shown in Algorithm~\ref{algorithm:Alg1}, successful executions start from a high base score, and penalties are subtracted according to completion latency, file I/O overhead, execution overhead, and recovery cost. Failed executions receive a lower base score determined by failure severity. We further distinguish delayed failures from immediate failures: if a task runs for a prolonged duration before failing, we add a small adjustment, as this indicates partial responsiveness under stress.

For successful tasks, the final score is primarily determined by execution efficiency. We model the completion-time penalty $p_t(\tau)$ as a piecewise linear function to capture different latency sensitivity regions. For failed executions, we assign a lower base score according to the error type and optionally add a delayed-failure adjustment when prolonged runtime indicates partial responsiveness.
Here, $p_t(\tau)$ denotes the completion-time penalty, $p_r(\tau)$ and $p_e(\tau)$ denote the penalties for file I/O overhead and model execution overhead respectively, and $p_o(\tau)$ is a recovery penalty determined by the recovery result $o(\tau)$.
The completion-time penalty $p_t(\tau)$ is defined as:
\begin{equation}
\label{pt}
p_t(\tau) = \beta_i + \alpha_i \cdot (t_s(\tau)-T_{i-1}), \quad  t_s(\tau) \in (T_{i-1}, T_i]
\end{equation}
where time thresholds $\{T_i\}$ and slopes $\alpha_i$ are empirical hyperparameters calibrated to reflect distinct sensitivity regions, and $\beta_i$ is the base penalty for the $i$-th stage.

\begin{algorithm}[h]
    \caption{Task Execution Score Calculation}
    \label{algorithm:Alg1}
    \begin{algorithmic}[1]
        \REQUIRE Inference task $\tau$, completion status $\tau_c$, completion time $t_s(\tau)$, source file reading time $t_r(\tau)$, model execution time $t_e(\tau)$, recovery result $o(\tau)$, error type $e(\tau)$
        \ENSURE Task execution score $S(\tau)$
        \IF{$\tau_c = 0$}
            \STATE Set basic score $R(\tau)$ based on $e(\tau)$
            \STATE $S(\tau) \leftarrow R(\tau)$
            \IF{$t_s(\tau) > 20$}
                \STATE Adjustment $a_f(\tau) \leftarrow \min\left(\frac{t_s(\tau)-20}{10}, 10\right)$
                \STATE $S(\tau) \leftarrow S(\tau) + a_f(\tau)$
            \ENDIF
        \ELSE
            \STATE Basic score $R(\tau) \leftarrow 100$
            \STATE Calculate time penalty $p_t(\tau)$ via Eq.~\ref{pt}
            \STATE $p_r(\tau) \leftarrow \min(\max(0, t_r(\tau) - 10)/2, 5)$
            \STATE $p_e(\tau) \leftarrow \min(\max(0, t_e(\tau) - 10)/2, 5)$
            \STATE Calculate recovery penalty $p_o(\tau)$ from $o(\tau)$
            \STATE $S(\tau) \leftarrow R(\tau) - \bigl(p_t(\tau) + p_r(\tau) + p_e(\tau) + p_o(\tau)\bigr)$
        \ENDIF
        \RETURN $S(\tau)$
    \end{algorithmic}
\end{algorithm}

\textbf{Heterogeneity-Aware Orchestrator.}
To handle the heterogeneity of training data, we design a length-aware workload scheduler. We first sort the dataset in descending order of prompt length and divide it into multiple batches. Then, based on each server's compute capability, partitions with longer contexts are assigned to high-performance servers, while those with shorter contexts go to low-power servers. This strategy alleviates synchronization issues in distributed fine-tuning. It also naturally clusters inputs of similar lengths, providing more samples of the same problem with different profiles to support the contrastive regularization in the loss computation. This strategy remains fully compatible with existing hardware-centric scheduling methods.

\textbf{Contrastive Regularization.} To prevent the model's score predictions from relying too heavily on the static content of the input, we propose a state-sensitivity loss. We penalize the similarity of the model's predictions for different structured profiles associated with the same target problem (or task). We use the term \textit{problem} to denote the target $q$ being evaluated in either scenario.

We first group the samples within each training batch by problem ID. For batch $\mathcal{B}$, we identify all samples sharing the same problem $q$ and index them into a set $G_q = \{ i \in \mathcal{B} \mid q_i = q \}$. The predicted logits $z_i$ for these samples are used for loss calculations. Since different subject states solving the same problem may yield distinct prediction patterns, we minimize the squared cosine similarity between their logits:
\begin{equation}
L_{\text{c}} = \frac{1}{\mathcal{N}} \sum_{q} \sum_{(i,j) \in G_q, i < j} \mathcal{S}^2(z_i, z_j)
\end{equation}
where $\mathcal{N}$ is the total number of valid pairs across all problem groups and $\mathcal{S}(z_i, z_j)$ denotes the cosine similarity between logit vectors $z_i$ and $z_j$:
\begin{equation}
\mathcal{S}(z_i, z_j) = \frac{z_i \cdot z_j}{\|z_i\| \cdot \|z_j\|}
\end{equation}

Minimizing $L_{\text{c}}$ encourages the model to produce different representations for different states, preventing it from collapsing into a problem-only shortcut. Although different subject profiles may ultimately yield the same performance score, this soft penalty ensures that the model arrives at its prediction by genuinely interpreting the distinct profile features rather than merely memorizing the problem semantics. >
Unlike standard contrastive learning that enforces invariance on positive pairs, we only apply a push mechanism on negative pairs, since enforcing similarity between the same profile solving different problems is not meaningful. The final fine-tuning objective is $L = L_{\text{ce}} + \lambda \cdot L_{\text{c}}$, where $L_{\text{ce}}$ is the classification cross-entropy loss and $\lambda$ balances the state-sensitivity penalty. The cross-entropy term dominates the actual score prediction, while the contrastive term acts as a regularizer to increase discrimination across different profile states.

\begin{table*}[t]
\caption{An ablation study of each component of our method.}
\label{Ablation}
\centering
\begin{tabular}{l c c c c c c c c c}
\toprule
\textbf{Methods}            & \textbf{Training Time (h)} & \textbf{Summary Sensitivity (\%)} & \textbf{Val} & \textbf{ACC} & \textbf{MAE} & \textbf{Precision} & \textbf{Recall} & \textbf{F1} & \textbf{Error Ratio (\%)} \\
\midrule
\multirow{2}{*}{FLASHBACK}  & \multirow{2}{*}{32.23}     & \multirow{2}{*}{33.2 ($K_G$)}     & $v_{i}$      & 0.68         & 13.4         & 0.69               & 0.63            & 0.68        & 5.6                       \\
                            &                            &                                   & $v_{f}$      & 0.63         & 16.1         & 0.65               & 0.63            & 0.64        & 7.3                       \\
\midrule
\multirow{2}{*}{EIP + Ray}  & \multirow{2}{*}{5.85}      & \multirow{2}{*}{40.9 ($E_G$)}     & $v_{i}$      & 0.61         & 15.7         & 0.62               & 0.61            & 0.62        & 4.7                       \\
                            &                            &                                   & $v_{f}$      & 0.34         & 25.7         & 0.39               & 0.34            & 0.32        & 11.0                      \\
\midrule
\multirow{2}{*}{Ray}        & \multirow{2}{*}{9.00}      & \multirow{2}{*}{37.9 ($K_G$)}     & $v_{i}$      & 0.68         & 12.8         & 0.69               & 0.68            & 0.69        & 4.6                       \\
                            &                            &                                   & $v_{f}$      & 0.64         & 14.7         & 0.66               & 0.64            & 0.65        & 5.7                       \\
\midrule
\multirow{2}{*}{PRay}       & \multirow{2}{*}{7.35}      & \multirow{2}{*}{36.0 ($K_G$)}     & $v_{i}$      & 0.66         & 13.7         & 0.67               & 0.66            & 0.66        & 5.0                       \\
                            &                            &                                   & $v_{f}$      & 0.64         & 15.3         & 0.64               & 0.64            & 0.64        & 6.1                       \\
\midrule
\multirow{2}{*}{CRay}       & \multirow{2}{*}{8.92}      & \multirow{2}{*}{29.4 ($K_G$)}     & $v_{i}$      & 0.65         & 14.5         & 0.65               & 0.65            & 0.65        & 5.2                       \\
                            &                            &                                   & $v_{f}$      & 0.58         & 17.6         & 0.59               & 0.58            & 0.58        & 6.1                       \\
\midrule
\multirow{2}{*}{Our}        & \multirow{2}{*}{\textbf{7.32}} & \multirow{2}{*}{\textbf{36.2} ($K_G$)} & $v_{i}$  & 0.66         & \textbf{13.5} & \textbf{0.67}      & \textbf{0.66}   & 0.67        & 4.8                       \\
                            &                            &                                   & $v_{f}$      & \textbf{0.64} & 16.2         & 0.65               & \textbf{0.64}   & 0.64        & 6.9                       \\
\bottomrule
\end{tabular}
\end{table*}

\begin{table}[t]
\caption{Training server and software settings in CogGuard.}
\label{setups}
\centering
\begin{tabular}{l l l}
\toprule
\textbf{Type} & \multicolumn{2}{l}{\textbf{Configuration}} \\
\midrule
\multirow{6}{*}{Hardware} & \multirow{5}{*}{x86} & CPU: Intel i9-14900KF @5.6GHz 24 Cores \\
 & & CPU: Intel Silver 4210 @2.20GHz 10 Cores \\
 & & CPU: Intel i9-10900K @3.70GHz 10 Cores \\
 & & GPU: 24GB 4090d, 24GB 3090, 8GB 2070s \\
\cmidrule(l){2-3}
 & ARM & Nvidia Spark (20 Cores, 128GB GB10 GPU) \\
\midrule
\multirow{2}{*}{Software} & \multicolumn{2}{l}{Ray: 2.52.1, nano-graphrag: 0.0.8.2,} \\
 & \multicolumn{2}{l}{Chaosd: v1.4, K8s: v1.26.10, LmDeploy: v0.6.4} \\
\bottomrule
\end{tabular}
\end{table}

\section{Experiments and Results}
\label{section_result}

\subsection{Experimental Setup}
\textbf{Datasets.} In the educational scenario, we evaluate mainstream SLMs across 40 students. In the operational scenario, we collect chaos injection and task execution results on mixed AI inference workloads from three heterogeneous servers \cite{zhang2025efficientmixedprecisionlargelanguage}. The dataset is split into training, validation, and test sets in an 8:1:1 ratio.

\textbf{Dataset Generalizability.} The datasets used in this study were obtained by crawling \textit{Nowcoder} and generating data via chaos injection and task testing on a heterogeneous edge cluster. The required data fields can also be obtained from other platforms such as \textit{PTA} and \textit{LeetCode}, making the data construction process transferable beyond the current testbed. Our data processing is limited to format conversion and the filtering of single-input multi-label data. Chaos injection and task testing procedures can be reproduced on any cluster. While we use \textit{LMDeploy} to deploy multimodal models for task execution, the execution tool can be replaced with \textit{vLLM}, \textit{Ollama}, or others.

\textbf{Setting.} The experimental infrastructure includes 4090, 4090D, 3090 workstations, and Spark edge devices. The distributed fine-tuning configurations are listed in Table \ref{setups}.
We construct a representative edge cluster consisting of a high-performance edge node (Nvidia Spark) and heterogeneous computing nodes (desktop hosts) to simulate the computational heterogeneity commonly encountered in edge environments.

\textbf{Experimental Parameters.} Table \ref{tab_model} lists the main experimental parameters. The comparison results under different contrastive weights are shown in Table \ref{tab_weight}.

\textbf{Baselines.} For Cognitive Profiling, we compare our prototype mapping approach with two baselines.
\begin{itemize}
	\item \textbf{Few-shot EI}\cite{wu2025embracingimperfectionsimulatingstudents}: Embracing Imperfection (EI) is a training-free method that constructs a static cognitive prototype for each student without graph clustering.
    \item \textbf{SimpleKT}\cite{liu2023simplekt}: This method uses Transformer to encode students' historical records and models problem-specific difficulty variations through Rasch decomposition, fusing these representations for score prediction.
\end{itemize}

For Alignment, we compare our method with five baselines.
\begin{itemize}
	\item \textbf{EIP}\cite{wu2025embracingimperfectionsimulatingstudents}: Embracing Imperfection Prediction (EIP) maps new tasks to this cognitive prototype to retrieve relevant concepts and predicts student performance using LLM-based inference without any parameter updates.
	\item \textbf{FLASHBACK}\cite{liu-etal-2025-flashback}: This method employs an appending context pattern that places retrieved documents at the end of the input. It is used to evaluate the time efficiency of distributed fine-tuning in Table \ref{Ablation}.
	\item \textbf{Ray}\cite{222605}: Ray is a widely used distributed task execution framework. We encapsulate fine-tuning programs into workers. Ray requires all servers to train with the same batch size and update weights through shared storage.
	\item \textbf{SLM-Probe}\cite{kumar-etal-2021-bert}: This method keeps all parameters of the Qwen2.5-3B model frozen and trains a lightweight 3-layer MLP classifier on top of the mean-pooled hidden states from the final layer.
	\item \textbf{GCN-Embed}\cite{NEURIPS2024_0b77d3a8}: This method converts entity and knowledge tokens into a token-level graph, which is then encoded by a 2-layer GCN followed by an MLP classifier.

\end{itemize}

\begin{table}[t]
\caption{Hyperparameter settings.}
\label{tab_model}
\centering
\begin{tabular}{l l l}
\toprule
\textbf{Type} & \textbf{Hyperparameter} & \textbf{Value} \\
\midrule
\multirow{6}{*}{Graph construction} & The Top-$k$ nodes in summary & 10 \\
 & Initial neutral score $\mu$ & 0.5 \\
 & LLM timeout & 60 s \\
 & LLM retry delay & 5 s \\
 & LLM max retries & 3 \\
 & LLM retry backoff & 2 \\
\midrule
\multirow{3}{*}{DDP} & num workers & 2 - 4 \\
 & per device train batch size & 4 \\
 & gradient accumulation steps & 6 \\
\midrule
\multirow{7}{*}{Fine-tuning} & train epochs & 48 \\
 & initial learning rate & 3e-5 \\
 & max chunk chars & 5000 \\
 & task type & SEQ\_CLS \\
 & contrastive weight & 0.1 \\
 & extra number of tokens & 4 \\
 & gradient checkpointing & TRUE \\
\bottomrule
\end{tabular}
\end{table}

\begin{table*}[t]
\centering
\caption{Performance of all methods across educational scenarios of different scales.}
\label{different_scales}
\scalebox{0.98}{
\begin{tabular}{l c c c c c c c c c c c c c c c}
\toprule
\multirow{2}{*}{\textbf{Methods}} & \multicolumn{5}{c}{\textbf{students\_20}} & \multicolumn{5}{c}{\textbf{students\_30}} & \multicolumn{5}{c}{\textbf{students\_40}} \\
\cmidrule(lr){2-6} \cmidrule(lr){7-11} \cmidrule(lr){12-16}
 & \textbf{ACC} & \textbf{MAE} & \textbf{Pre.} & \textbf{Recall} & \textbf{F1} & \textbf{ACC} & \textbf{MAE} & \textbf{Pre.} & \textbf{Recall} & \textbf{F1} & \textbf{ACC} & \textbf{MAE} & \textbf{Pre.} & \textbf{Recall} & \textbf{F1} \\
\midrule
Few-shot EI & 0.26 & 37.5 & 0.22 & 0.26 & 0.23 & 0.25 & 38.8 & 0.23 & 0.25 & 0.23 & 0.28 & 36.4 & 0.24 & 0.28 & 0.25 \\
SimpleKT & 0.59 & 15.3 & 0.64 & 0.59 & 0.61 & 0.59 & 16.2 & 0.62 & 0.59 & 0.60 & 0.61 & 16.9 & 0.62 & 0.61 & 0.61 \\
SLM-Probe & 0.37 & 32.3 & 0.38 & 0.37 & 0.23 & 0.50 & 20.1 & 0.51 & 0.50 & 0.49 & 0.50 & 21.1 & 0.50 & 0.50 & 0.49 \\
GCN-Embed & 0.50 & 21.4 & 0.52 & 0.50 & 0.51 & 0.50 & 23.6 & 0.50 & 0.50 & 0.49 & 0.49 & 23.0 & 0.49 & 0.49 & 0.48 \\
Our (Qwen2.5-3B) & 0.60 & \textbf{15.4} & 0.62 & 0.60 & 0.61 & 0.61 & \textbf{14.8} & 0.61 & 0.61 & 0.61 & 0.63 & \textbf{14.3} & 0.63 & 0.63 & 0.63 \\
\bottomrule
\end{tabular}}
\end{table*}

\begin{table*}[t]
\caption{Details of prediction results.}
\label{tab:results}
\centering
\scalebox{0.9}{
\begin{tabular}{c c c c c c c c c c}
\toprule
\textbf{Score Intervals} & \textbf{Number} & \textbf{Metrics} & \textbf{Qwen3-0.6B} & \textbf{Qwen3-1.7B} & \textbf{Gemma-2-2B-it} & \textbf{Qwen2.5-3B} & \textbf{Qwen3-8B} & \textbf{Llama-3.1-8B} & \textbf{Gemma-2-9B-it} \\
\midrule
\multirow{3}{*}{0-20}   & \multirow{3}{*}{234} & Precision & 0.74 & 0.72 & \textbf{0.75} & 0.72 & 0.74 & 0.73 & 0.74 \\
                        &                      & Recall    & 0.75 & \textbf{0.79} & 0.75 & 0.75 & \textbf{0.79} & 0.74 & 0.76 \\
                        &                      & F1        & 0.75 & 0.75 & 0.75 & 0.74 & \textbf{0.76} & 0.73 & 0.75 \\
\midrule
\multirow{3}{*}{20-40}  & \multirow{3}{*}{98}  & Precision & 0.60 & 0.60 & 0.61 & 0.55 & 0.62 & 0.58 & \textbf{0.65} \\
                        &                      & Recall    & 0.60 & 0.48 & 0.61 & 0.57 & \textbf{0.65} & 0.60 & 0.63 \\
                        &                      & F1        & 0.60 & 0.53 & 0.61 & 0.56 & 0.63 & 0.59 & \textbf{0.64} \\
\midrule
\multirow{3}{*}{40-60}  & \multirow{3}{*}{72}  & Precision & 0.40 & 0.41 & \textbf{0.51} & 0.37 & 0.44 & 0.42 & 0.45 \\
                        &                      & Recall    & 0.39 & 0.33 & \textbf{0.42} & 0.38 & 0.38 & 0.39 & 0.40 \\
                        &                      & F1        & 0.39 & 0.37 & \textbf{0.46} & 0.37 & 0.41 & 0.40 & 0.43 \\
\midrule
\multirow{3}{*}{60-80}  & \multirow{3}{*}{89}  & Precision & 0.58 & 0.58 & \textbf{0.61} & 0.55 & 0.56 & 0.60 & \textbf{0.61} \\
                        &                      & Recall    & 0.58 & 0.55 & \textbf{0.60} & 0.47 & 0.54 & 0.58 & \textbf{0.60} \\
                        &                      & F1        & 0.58 & 0.57 & \textbf{0.60} & 0.51 & 0.55 & 0.59 & \textbf{0.60} \\
\midrule
\multirow{3}{*}{80-100} & \multirow{3}{*}{116} & Precision & 0.66 & 0.58 & 0.68 & 0.64 & 0.62 & 0.63 & \textbf{0.69} \\
                        &                      & Recall    & 0.66 & 0.67 & \textbf{0.76} & 0.64 & 0.59 & 0.63 & 0.72 \\
                        &                      & F1        & 0.66 & 0.62 & \textbf{0.72} & 0.64 & 0.61 & 0.63 & 0.71 \\
\bottomrule
\end{tabular}
}
\end{table*}

\begin{table}[t]
\caption{Experimental results under different contrastive weights.}
\label{tab_weight}
\centering
\begin{tabular}{c c c c c}
\toprule
\textbf{Contrastive Weight} & \textbf{ACC} & \textbf{MAE} & \textbf{F1} & \textbf{Error Ratio (\%)} \\
\midrule
0.05 & 0.66 & 13.8 & 0.66 & 3.4 \\
0.1  & \textbf{0.66} & \textbf{13.9} & \textbf{0.66} & \textbf{2.6} \\
0.15 & 0.64 & 14.9 & 0.64 & 3.4 \\
0.2  & 0.65 & 14.2 & 0.64 & 2.6 \\
0.3  & 0.64 & 14.7 & 0.64 & 3.2 \\
\bottomrule
\end{tabular}
\end{table}

\begin{table}[t]
\caption{Performance of our method across different base models.}
\label{different_finetuning}
\centering
\begin{tabular}{l c c c c}
\toprule
\textbf{Models} & \textbf{ACC} & \textbf{MAE} & \textbf{Macro F1} & \textbf{Weighted F1} \\
\midrule
Qwen3-0.6B & 0.64 & 14.2 & 0.60 & 0.64 \\
Qwen3-1.7B & 0.63 & 14.8 & 0.57 & 0.62 \\
Gemma-2-2B-it & 0.67 & 13.5 & 0.63 & 0.67 \\
Qwen2.5-3B & 0.63 & 14.3 & 0.58 & 0.63 \\
Qwen3-8B & 0.64 & 14.0 & 0.59 & 0.64 \\
Llama-3.1-8B & 0.63 & 15.2 & 0.60 & 0.63 \\
Gemma-2-9B-it & \textbf{0.67} & \textbf{13.4} & \textbf{0.63} & \textbf{0.67} \\
\bottomrule
\end{tabular}
\end{table}

\subsection{Ablation Study}
We validate each component of our method through an ablation study, with results in Table \ref{Ablation}. Summary Sensitivity denotes the proportion of prediction changes when the model receives the same problem but randomly sampled profile summaries. Error Ratio denotes the proportion of cases where a score range of 0--20 is misclassified as 80--100, or vice versa.
We report experiments on the students\_20 subset in the educational scenario under identical conditions. PRay refers to the integration of Ray with training data partitioning, while CRay refers to the integration of Ray with contrastive regularization.

Because the profile summary is a lossy abstraction of the full submission record, different code variants may occasionally collapse to similar summaries while receiving different scores. To analyze how such ambiguity affects validation performance, we construct a filtered subset $v_{f}$ as a diagnostic evaluation while retaining the original validation split $v_{i}$ as the primary benchmark. Although PRay achieves competitive performance on $v_f$, our method further improves the primary benchmark $v_i$ and achieves the highest summary sensitivity. This indicates that contrastive regularization reduces problem-memorization shortcuts, forcing the model to rely more on personalized cognitive profiles in noisy, real-world settings.

\begin{figure*}[t]
  \centering
  \includegraphics[width=1\textwidth]{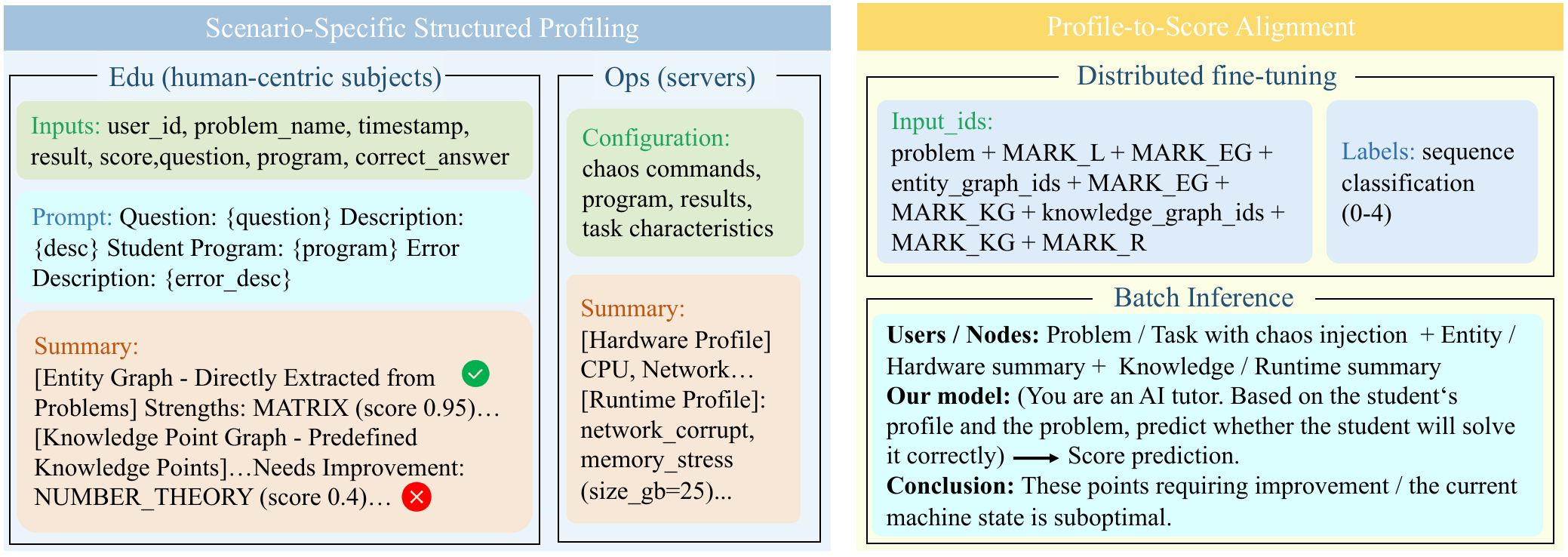}
  \centering
  \caption{Application workflow of CogGuard.}
  \label{fig:illustration}
\end{figure*}

\begin{figure}[t]
	\centering
	\includegraphics[width=1\linewidth]{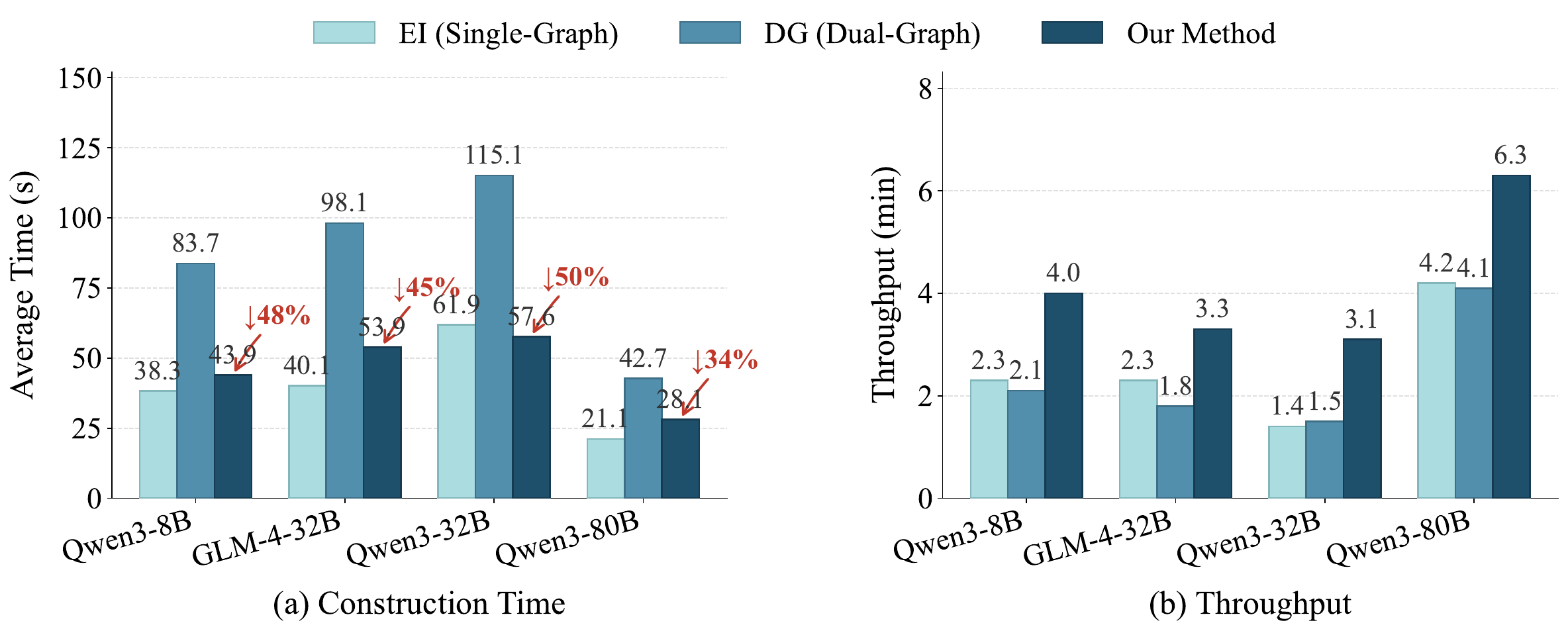}
	\centering
	\caption{Efficiency of structured profiling across models.}
	\label{fig:cost}
\end{figure}

\begin{table}[htbp]
    \centering
    \caption{Dynamic evolution of a student's cognitive profile.}
    \label{tab:profile_evolution}
    \begin{tabular}{@{} p{0.22\linewidth} p{0.33\linewidth} p{0.33\linewidth} @{}}
        \toprule
        \textbf{Attribute} & \textbf{Record 62 (Attempt 1)} & \textbf{Record 63 (Attempt 2)} \\
        \midrule

        Problem & \multicolumn{2}{p{0.68\linewidth}}{``Maximize the sum of a sub-rectangle in an $n \times m$ matrix generated by two weight arrays.''} \\
        \midrule

        Score & 50 & 0 \\
        \midrule

        Stable \newline Strengths & \multicolumn{2}{p{0.68\linewidth}}{Greedy (0.95), Control Structure (0.95), Input Output and Sequential Structure(0.95), Search (0.95), Simulation (0.95), Enumeration and Sorting (0.95), Number Theory (0.95), Hash (0.95), Dynamic Programming (0.95)} \\
        \midrule

        Stable \newline Weaknesses & \multicolumn{2}{p{0.68\linewidth}}{Problem Requirements (0.35), Error Description (0.35), Bitmasking (0.35), Linked List (0.40)} \\
        \midrule

        Dynamic \newline Shifts & {\textbf{Graph Structure (0.59)}} & {\textbf{Probability \& Statistics (0.60)}} \\

        \bottomrule
    \end{tabular}
\end{table}

\begin{figure}[t]
	\centering
	\includegraphics[width=1\linewidth]{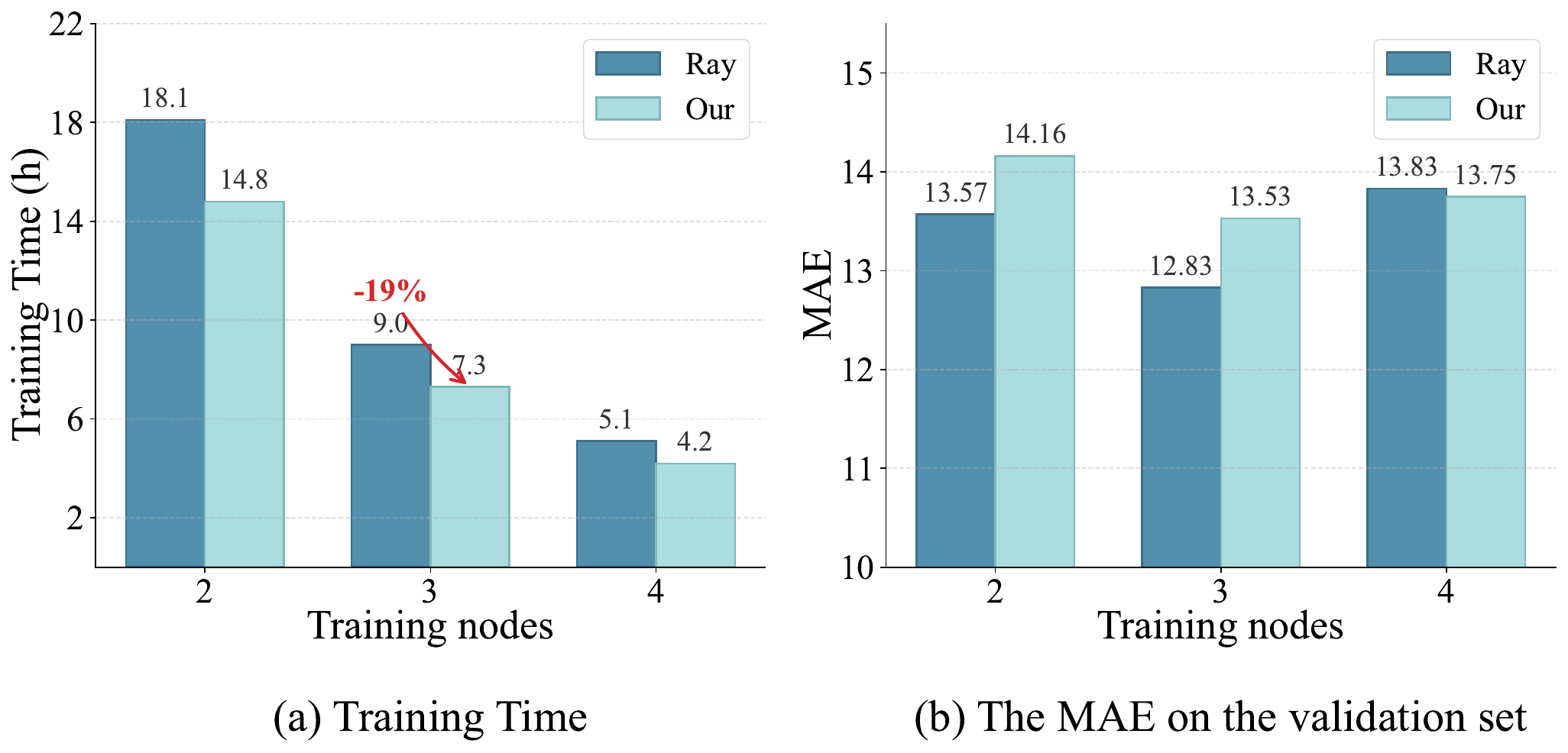}
	\centering
	\caption{Performance on heterogeneous clusters.}
	\label{fig:heterogeneous}
\end{figure}

\subsection{Effectiveness of Profile-to-Score Alignment}
Table \ref{different_scales} presents the performance of all methods across educational datasets of different scales. SLM-Probe shows a strong dependency on dataset size.
Few-shot EI yields poor retrieval performance because it lacks graph clustering support. In addition, increasing the dataset size consistently improves the performance of our method.

In Table \ref{different_finetuning}, we compare the performance of our method across different fine-tuning base models, using the cleaned students\_40 dataset for testing. Our method achieves performance gains as the model size increases, with the Gemma model showing better stability. The minimum Mean Absolute Error (MAE) is approximately 13 on a 100-point-scale. While an MAE of 13.4 still indicates noticeable prediction variance, it remains useful for early intervention because it separates high-risk failures from safe completions without requiring heavy cloud-side computation.

We detail the prediction results of our method in different score ranges in Table \ref{tab:results}, where the label distribution in the validation set is consistent with that of the training set. Performance degrades in score intervals with fewer samples. However, given that the training data is derived from real-world scenarios, this distribution reflects the actual scoring patterns of C++ programs. Despite the scarcity of training samples in the intermediate score intervals, the Gemma model still achieves the best performance. In contrast, the Qwen model tends to achieve better predictions in score intervals with more abundant training samples. In future work, we plan to address this issue by generating additional realistic samples.

\subsection{Efficiency of Dual-Graph Structured Profiling}

We use high-capacity LLMs for offline profile extraction and fine-tuned SLMs for real-time edge-side alignment, matching the different latency and reasoning requirements of the two stages.
In Fig. \ref{fig:cost}, we evaluate the efficiency of dual-graph cognitive profiling. Since the model Qwen3-Next-80B does not support prefix-caching, our method achieves a 34.2\% reduction compared with dual-graph (DG) construction on this model. With the combined optimization of prompt design and KV cache reuse, we obtain a 47.7\% reduction in average time on the other models, which is close to the single-graph construction overhead of the EI method. This shows that the additional time cost of dual-graph profiling is small. Cognitive profiling can also run entirely on edge devices with a slight drop in precision, as shown by Qwen3-8B.

To illustrate the dynamic tracking capability of our method, Table \ref{tab:profile_evolution} details the evolution of a student's cognitive profile across two consecutive attempts on a specific array manipulation problem. The system captures the shifting cognitive focus from ``Graph Structure'' to ``Probability and Statistics'', correlating with the score drop from 50 to 0.

\begin{figure}[t]
	\centering
	\includegraphics[width=1\linewidth]{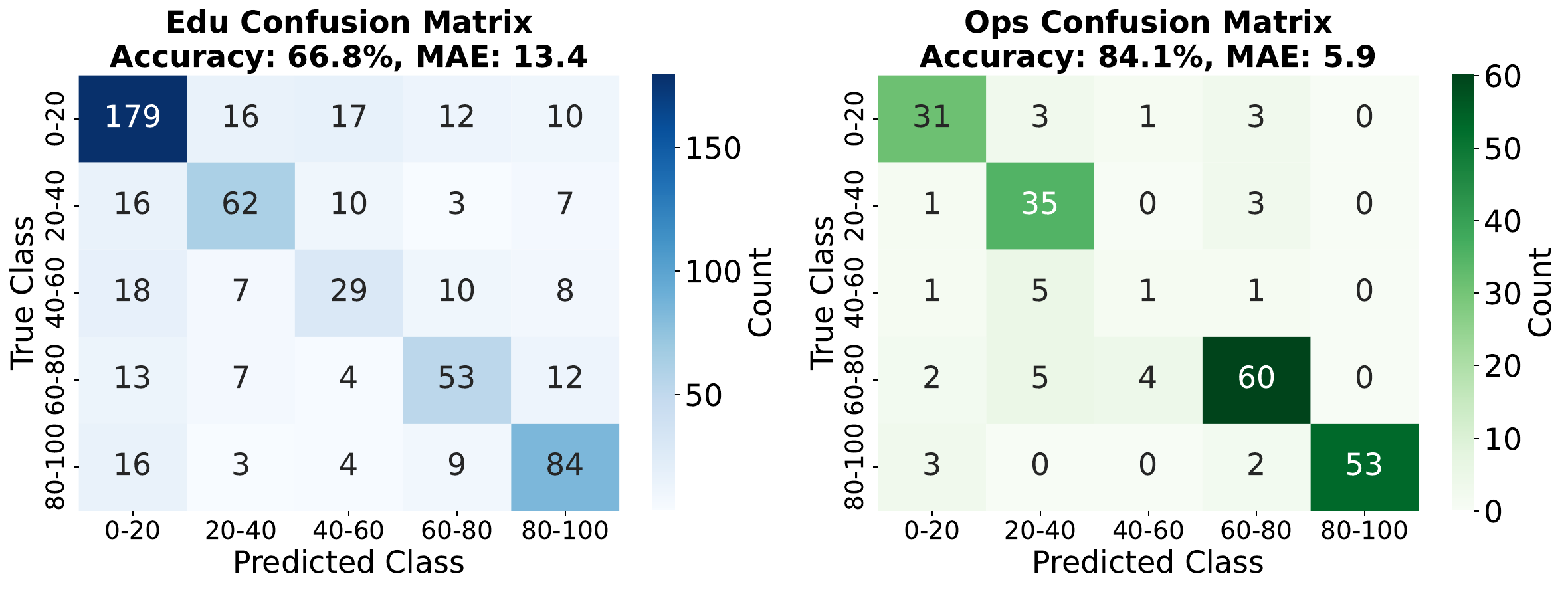}
	\centering
	\caption{Confusion matrix results of our method in cross-scenario settings.}
	\label{fig:confusion}
\end{figure}

\begin{figure}[t]
	\centering
	\includegraphics[width=1\linewidth]{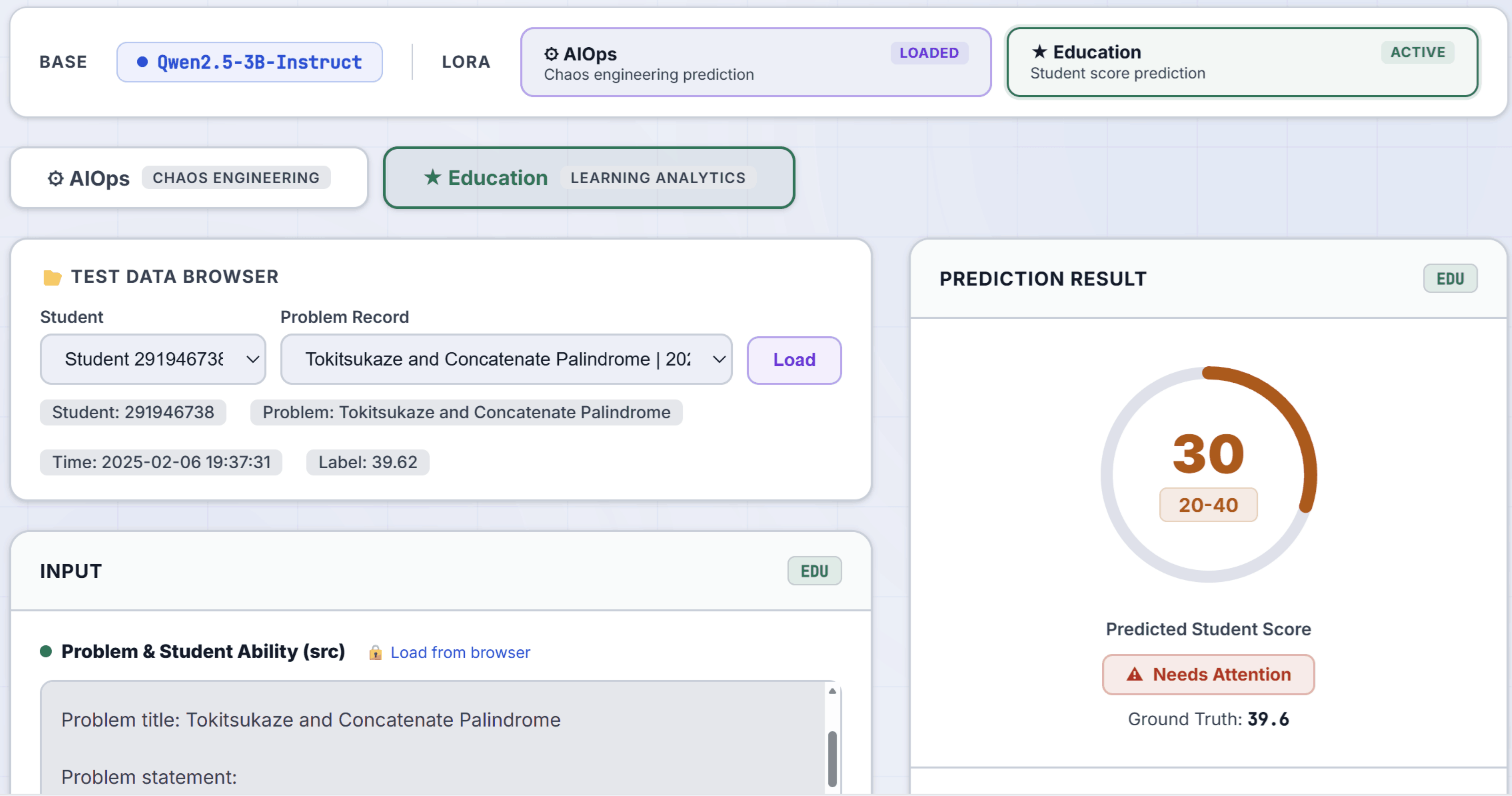}
	\centering
	\caption{End-to-end service execution pipeline of CogGuard.}
	\label{fig:service_execution_pipeline}
\end{figure}

\subsection{Performance on Heterogeneous Clusters}
In Fig. \ref{fig:heterogeneous}, we show the performance of our method on heterogeneous clusters. The server settings correspond to (3090 + Spark), (4090 + 3090 + Spark), and (4090 + 4090D + 3090 + Spark), respectively.
The results show that training data partitioning has minimal impact on prediction accuracy, while our method achieves an additional time reduction of nearly 19\% on top of Ray-based distributed training. This training overhead is far smaller than the time typically required for graph clustering and community report generation in Graph RAG \cite{zhao2025e2graphragstreamlininggraphbasedrag,han2026ragvsgraphragsystematic}, confirming that our method adds very little time cost.

\subsection{Cross-Scenario Experimental Results}
We further evaluate CogGuard in the operational scenario, where the model achieves lower prediction error than in the educational scenario. This is expected because operational faults are artificially injected with deterministic physical boundaries, whereas human behavioral features contain more noise and individual variation. Nevertheless, the results in both scenarios show that the proposed alignment pipeline can handle both machine-state profiles and human-centered cognitive profiles. Fig. \ref{fig:illustration} provides an overview of the proposed workflow, and the confusion matrices in Fig. \ref{fig:confusion} further summarize our score prediction performance. We illustrate the end-to-end service execution pipeline in Fig. \ref{fig:service_execution_pipeline}. The system retrieves pre-stored profiles and target tasks for designated subjects to deliver batch proactive warning services.

\subsection{Limitations}
The educational profile summary is a lossy abstraction that may miss fine-grained code-level distinctions, while the operational dataset is derived from controlled chaos injection and may not fully reflect organic production failures. In addition, our cross-scenario study mainly validates a shared alignment interface rather than a fully unified upstream profiling process. Since this proactive-warning setting is not yet standardized, fully task-matched baselines are currently unavailable. We also note that the current profiling mechanism has minor limitations, such as formatting artifacts during LLM extraction that occasionally yield redundant knowledge entries; we plan to improve this extraction robustness in future work.

\section{Conclusion}
\label{section_conclusion}
This paper presented CogGuard, a proactive-warning method for edge intelligent services that combines scenario-specific profiling with a shared profile-to-score alignment pipeline. We formulated proactive warning as a static-dynamic profile-to-score prediction problem and proposed a dual-graph cognitive profiling mechanism with prefix-aligned KV cache reuse for educational scenarios and a chaos-based profiling approach for operational scenarios. To support efficient edge deployment, we introduced a length-aware distributed fine-tuning strategy with contrastive regularization to reduce synchronization bottlenecks and prevent problem-memorization shortcuts. Experiments on both educational and operational datasets showed that CogGuard reduced profile construction and training time by 48\% and 19\%, respectively, while achieving MAEs of 13.4 and 5.9 on 100-point-scale warning tasks. Future work includes improving the robustness of LLM-based entity extraction to reduce formatting artifacts, exploring data augmentation for underrepresented score intervals, and extending the profiling pipeline to additional edge service scenarios such as predictive maintenance and personalized recommendation.



\bibliographystyle{IEEEtran}
\bibliography{reference}

@article{cite-key,
	author = {Wang, Sili and Yang, Heng and Liu, Wei},
	doi = {10.1038/s41598-025-21222-z},
	journal = {Scientific Reports},
	number = {1},
	pages = {40425},
	title = {Research on the construction and application of retrieval enhanced generation (RAG) model based on knowledge graph},
	volume = {15},
	year = {2025}
}

@inproceedings{wu-etal-2025-medical,
    title = "Medical Graph {RAG}: Evidence-based Medical Large Language Model via Graph Retrieval-Augmented Generation",
    author = "Wu, Junde  and
      Zhu, Jiayuan  and
      Qi, Yunli  and
      Chen, Jingkun  and
      Xu, Min  and
      Menolascina, Filippo  and
      Jin, Yueming  and
      Grau, Vicente",
    booktitle = "Proceedings of the 63rd Annual Meeting of the Association for Computational Linguistics",
    month = jul,
    year = "2025",
    address = "Vienna, Austria",
    publisher = "Association for Computational Linguistics",
    doi = "10.18653/v1/2025.acl-long.1381",
    pages = "28443--28467"
}

@article{MIAO2026114855,
	author = {Runsheng Miao and Tao Wu and Zhenyu Zhang},
	doi = {https://doi.org/10.1016/j.knosys.2025.114855},
	journal = {Knowledge-Based Systems},
	pages = {114855},
	title = {Graph RAG-based fault diagnosis for train bogies using knowledge graphs and large language model},
	volume = {331},
	year = {2026}
}

@inproceedings{mavromatis-karypis-2025-gnn,
    title = "{GNN}-{RAG}: Graph Neural Retrieval for Efficient Large Language Model Reasoning on Knowledge Graphs",
    author = "Mavromatis, Costas  and
      Karypis, George",
    booktitle = "Findings of the Association for Computational Linguistics: ACL 2025",
    month = jul,
    year = "2025",
    address = "Vienna, Austria",
    publisher = "Association for Computational Linguistics",
    doi = "10.18653/v1/2025.findings-acl.856",
    pages = "16682--16699"
}

@inproceedings{piech2015deep,
  title={Deep knowledge tracing},
  author={Piech, Chris and Bassen, Jonathan and Huang, Jonathan and Ganguli, Surya and Sahami, Mehran and Guibas, Leonidas and Sohl-Dickstein, Jascha},
  booktitle={Proceedings of the Advances in Neural Information Processing Systems},
  volume={28},
  year={2015}
}

@misc{wu2025embracingimperfectionsimulatingstudents,
      title={Embracing Imperfection: Simulating Students with Diverse Cognitive Levels Using LLM-based Agents}, 
      author={Tao Wu and Jingyuan Chen and Wang Lin and Mengze Li and Yumeng Zhu and Ang Li and Kun Kuang and Fei Wu},
      year={2025},
      eprint={2505.19997},
      archivePrefix={arXiv},
      primaryClass={cs.LG},
      howpublished={arXiv preprint arXiv:2505.19997}
}

@inproceedings{Lv_Liu_Gao_Zhang_Lu_Zhu_2025,
	author = {Lv, Rui and Liu, Qi and Gao, Weibo and Zhang, Haotian and Lu, Junyu and Zhu, Linbo},
	doi = {10.1609/aaai.v39i1.32038},
	booktitle = {Proceedings of the AAAI Conference on Artificial Intelligence},
	month = {Apr.},
	pages = {577-585},
	title = {GenAL: Generative Agent for Adaptive Learning},
	volume = {39},
	year = {2025}
}

@misc{liu2025dualreasoninggnnllmcollaborative,
      title={Dual Reasoning: A GNN-LLM Collaborative Framework for Knowledge Graph Question Answering}, 
      author={Guangyi Liu and Yongqi Zhang and Yong Li and Quanming Yao},
      year={2025},
      eprint={2406.01145},
      archivePrefix={arXiv},
      primaryClass={cs.CL},
      howpublished={arXiv preprint arXiv:2406.01145}
}

@misc{han2025retrievalaugmentedgenerationgraphsgraphrag,
      title={Retrieval-Augmented Generation with Graphs (GraphRAG)}, 
      author={Haoyu Han and Yu Wang and Harry Shomer and Kai Guo and Jiayuan Ding and Yongjia Lei and Mahantesh Halappanavar and Ryan A. Rossi and Subhabrata Mukherjee and Xianfeng Tang and Qi He and Zhigang Hua and Bo Long and Tong Zhao and Neil Shah and Amin Javari and Yinglong Xia and Jiliang Tang},
      year={2025},
      eprint={2501.00309},
      archivePrefix={arXiv},
      primaryClass={cs.IR},

      howpublished={arXiv preprint arXiv:2501.00309}
}

@inproceedings {222605,
    author = {Philipp Moritz and Robert Nishihara and Stephanie Wang and Alexey Tumanov and Richard Liaw and Eric Liang and Melih Elibol and Zongheng Yang and William Paul and Michael I. Jordan and Ion Stoica},
    title = {Ray: A Distributed Framework for Emerging {AI} Applications},
    booktitle = {Proceedings of the 13th USENIX Symposium on Operating Systems Design and Implementation (OSDI 18)},
    year = {2018},
    address = {Carlsbad, CA},
    pages = {561--577},
    publisher = {USENIX Association},
    month = oct
}

@inproceedings {254418,
    author = {Jay H. Park and Gyeongchan Yun and Chang M. Yi and Nguyen T. Nguyen and Seungmin Lee and Jaesik Choi and Sam H. Noh and Young-ri Choi},
    title = {{HetPipe}: Enabling Large {DNN} Training on (Whimpy) Heterogeneous {GPU} Clusters through Integration of Pipelined Model Parallelism and Data Parallelism},
    booktitle = {Proceedings of the 2020 USENIX Annual Technical Conference (USENIX ATC 20)},
    year = {2020},
    pages = {307--321},
    publisher = {USENIX Association},
    month = jul
}

@INPROCEEDINGS{9065574,
  author={Song, Linghao and Chen, Fan and Zhuo, Youwei and Qian, Xuehai and Li, Hai and Chen, Yiran},
  booktitle={Proceedings of the 2020 IEEE International Symposium on High Performance Computer Architecture (HPCA)}, 
  title={AccPar: Tensor Partitioning for Heterogeneous Deep Learning Accelerators}, 
  year={2020},
  volume={},
  number={},
  pages={342-355}
}

@inproceedings{liu2023simplekt,
    title={simple{KT}: A Simple But Tough-to-Beat Baseline for Knowledge Tracing},
    author={Zitao Liu and Qiongqiong Liu and Jiahao Chen and Shuyan Huang and Weiqi Luo},
    booktitle={Proceedings of the Eleventh International Conference on Learning Representations},
    year={2023}
}

@inproceedings{NEURIPS2023_8cf04c64,
	author = {Hu, Liya and Dong, Zhiang and Chen, Jingyuan and Wang, Guifeng and Wang, Zhihua and Zhao, Zhou and Wu, Fei},
	booktitle = {Proceedings of the Advances in Neural Information Processing Systems},
	pages = {44976--44996},
	publisher = {Curran Associates, Inc.},
	title = {PTADisc: A Cross-Course Dataset Supporting Personalized Learning in Cold-Start Scenarios},
	volume = {36},
	year = {2023}
}

@misc{yang2025codethinkthinkcode,
      title={Code to Think, Think to Code: A Survey on Code-Enhanced Reasoning and Reasoning-Driven Code Intelligence in LLMs}, 
      author={Dayu Yang and Tianyang Liu and Daoan Zhang and Antoine Simoulin and Xiaoyi Liu and Yuwei Cao and Zhaopu Teng and Xin Qian and Grey Yang and Jiebo Luo and Julian McAuley},
      year={2025},
      eprint={2502.19411},
      archivePrefix={arXiv},
      primaryClass={cs.CL},
      howpublished={arXiv preprint arXiv:2502.19411}
}

@INPROCEEDINGS{10903891,
  author={Mailewa, Akalanka B. and Akuthota, Arunkumar and Mohottalalage, Thivanka M. Dissanayake},
  booktitle={Proceedings of the 2025 IEEE 15th Annual Computing and Communication Workshop and Conference (CCWC)}, 
  title={A Review of Resilience Testing in Microservices Architectures: Implementing Chaos Engineering for Fault Tolerance and System Reliability}, 
  year={2025},
  volume={},
  number={},
  pages={00236-00242},
  doi={10.1109/CCWC62904.2025.10903891}}

@misc{yang2025graphusionragframeworkknowledge,
      title={Graphusion: A RAG Framework for Knowledge Graph Construction with a Global Perspective}, 
      author={Rui Yang and Boming Yang and Aosong Feng and Sixun Ouyang and Moritz Blum and Tianwei She and Yuang Jiang and Freddy Lecue and Jinghui Lu and Irene Li},
      year={2025},
      eprint={2410.17600},
      archivePrefix={arXiv},
      primaryClass={cs.CL},
      howpublished={arXiv preprint arXiv:2410.17600}
}

@misc{zhang2025leanragknowledgegraphbasedgenerationsemantic,
      title={LeanRAG: Knowledge-Graph-Based Generation with Semantic Aggregation and Hierarchical Retrieval}, 
      author={Yaoze Zhang and Rong Wu and Pinlong Cai and Xiaoman Wang and Guohang Yan and Song Mao and Ding Wang and Botian Shi},
      year={2025},
      eprint={2508.10391},
      archivePrefix={arXiv},
      primaryClass={cs.AI},
      howpublished={arXiv preprint arXiv:2508.10391}
}

@misc{edge2025localglobalgraphrag,
      title={From Local to Global: A Graph RAG Approach to Query-Focused Summarization}, 
      author={Darren Edge and Ha Trinh and Newman Cheng and Joshua Bradley and Alex Chao and Apurva Mody and Steven Truitt and Dasha Metropolitansky and Robert Osazuwa Ness and Jonathan Larson},
      year={2025},
      eprint={2404.16130},
      archivePrefix={arXiv},
      primaryClass={cs.CL},
      howpublished={arXiv preprint arXiv:2404.16130}
}

@inproceedings{NEURIPS2024_efaf1c97,
	author = {He, Xiaoxin and Tian, Yijun and Sun, Yifei and Chawla, Nitesh V. and Laurent, Thomas and LeCun, Yann and Bresson, Xavier and Hooi, Bryan},
	booktitle = {Proceedings of the Advances in Neural Information Processing Systems},
	doi = {10.52202/079017-4224},
	pages = {132876--132907},
	publisher = {Curran Associates, Inc.},
	title = {G-Retriever: Retrieval-Augmented Generation for Textual Graph Understanding and Question Answering},
	volume = {37},
	year = {2024}
}

@article{SEGAL2019261,
	author = {Avi Segal and Kobi Gal and Guy Shani and Bracha Shapira},
	doi = {https://doi.org/10.1016/j.ijhcs.2019.07.002},
	journal = {International Journal of Human-Computer Studies},
	pages = {261-272},
	title = {A difficulty ranking approach to personalization in E-learning},
	volume = {130},
	year = {2019}
}

@misc{zhang2025efficientmixedprecisionlargelanguage,
      title={Efficient Mixed-Precision Large Language Model Inference with TurboMind}, 
      author={Li Zhang and Youhe Jiang and Guoliang He and Xin Chen and Han Lv and Qian Yao and Fangcheng Fu and Kai Chen},
      year={2025},
      eprint={2508.15601},
      archivePrefix={arXiv},
      primaryClass={cs.DC},
      howpublished={arXiv preprint arXiv:2508.15601}
}

@article{10.1145/3777375,
    author = {Owotogbe, Joshua and Kumara, Indika and Heuvel, Willem-Jan and Tamburri, Damian},
    title = {Chaos Engineering: A Multi-Vocal Literature Review},
    year = {2025},
    issue_date = {May 2026},
    publisher = {Association for Computing Machinery},
    address = {New York, NY, USA},
    volume = {58},
    number = {7},
    doi = {10.1145/3777375},
    journal = {ACM Comput. Surv.},
    month = dec,
    articleno = {164},
    numpages = {44}
}

@inproceedings{kumar-etal-2021-bert,
    title = "What {BERT} Based Language Model Learns in Spoken Transcripts: An Empirical Study",
    author = "Kumar, Ayush  and
      Narayanan Sundararaman, Mukuntha  and
      Vepa, Jithendra",
    booktitle = "Proceedings of the Fourth BlackboxNLP Workshop on Analyzing and Interpreting Neural Networks for NLP",
    month = nov,
    year = "2021",
    address = "Punta Cana, Dominican Republic",
    publisher = "Association for Computational Linguistics",
    doi = "10.18653/v1/2021.blackboxnlp-1.25",
    pages = "322--336"
}

@inproceedings{NEURIPS2024_0b77d3a8,
	author = {Wang, Duo and Zuo, Yuan and Li, Fengzhi and Wu, Junjie},
	booktitle = {Proceedings of the Advances in Neural Information Processing Systems},
	doi = {10.52202/079017-0193},
	pages = {5950--5973},
	publisher = {Curran Associates, Inc.},
	title = {LLMs as Zero-shot Graph Learners: Alignment of GNN Representations with LLM Token Embeddings},
	volume = {37},
	year = {2024}
}

@inproceedings{liu-etal-2025-flashback,
    title = "{F}lash{B}ack: Efficient Retrieval-Augmented Language Modeling for Fast Inference",
    author = "Liu, Runheng  and
      Xiao, Xingchen  and
      Huang, Heyan  and
      Chi, Zewen  and
      Wu, Zhijing",
    booktitle = "Findings of the Association for Computational Linguistics: ACL 2025",
    month = jul,
    year = "2025",
    address = "Vienna, Austria",
    publisher = "Association for Computational Linguistics",
    doi = "10.18653/v1/2025.findings-acl.33",
    pages = "595--608"
}

@inproceedings{10.1145/3589334.3645574,
    author = {Wu, Songhao and Tu, Quan and Liu, Hong and Xu, Jia and Liu, Zhongyi and Zhang, Guannan and Wang, Ran and Chen, Xiuying and Yan, Rui},
    title = {Unify Graph Learning with Text: Unleashing LLM Potentials for Session Search},
    year = {2024},
    publisher = {Association for Computing Machinery},
    address = {New York, NY, USA},
    doi = {10.1145/3589334.3645574},
    booktitle = {Proceedings of the ACM Web Conference 2024},
    pages = {1509–1518},
    numpages = {10},
    location = {Singapore, Singapore},
}

@inproceedings{10.1145/3637528.3671810,
author = {Zhong, Aoxiao and Mo, Dengyao and Liu, Guiyang and Liu, Jinbu and Lu, Qingda and Zhou, Qi and Wu, Jiesheng and Li, Quanzheng and Wen, Qingsong},
title = {LogParser-LLM: Advancing Efficient Log Parsing with Large Language Models},
year = {2024},
publisher = {Association for Computing Machinery},
address = {New York, NY, USA},
doi = {10.1145/3637528.3671810},
booktitle = {Proceedings of the 30th ACM SIGKDD Conference on Knowledge Discovery and Data Mining},
pages = {4559–4570},
numpages = {12},
location = {Barcelona, Spain}
}

@ARTICLE{10835069,
  author={Qu, Guanqiao and Chen, Qiyuan and Wei, Wei and Lin, Zheng and Chen, Xianhao and Huang, Kaibin},
  journal={IEEE Communications Surveys \& Tutorials}, 
  title={Mobile Edge Intelligence for Large Language Models: A Contemporary Survey}, 
  year={2025},
  volume={27},
  number={6},
  pages={3820-3860},
  doi={10.1109/COMST.2025.3527641}
}

@inproceedings{srivatsa2025preble,
    title={Preble: Efficient Distributed Prompt Scheduling for {LLM} Serving},
    author={Vikranth Srivatsa and Zijian He and Reyna Abhyankar and Dongming Li and Yiying Zhang},
    booktitle={Proceedings of the Thirteenth International Conference on Learning Representations},
    year={2025},
}

@misc{zhong2025semragsemanticknowledgeaugmentedrag,
      title={SemRAG: Semantic Knowledge-Augmented RAG for Improved Question-Answering}, 
      author={Kezhen Zhong and Basem Suleiman and Abdelkarim Erradi and Shijing Chen},
      year={2025},
      eprint={2507.21110},
      archivePrefix={arXiv},
      primaryClass={cs.CL},
      howpublished={arXiv preprint arXiv:2507.21110}
}

@article{10.1145/3719664,
    author = {Zheng, Yue and Chen, Yuhao and Qian, Bin and Shi, Xiufang and Shu, Yuanchao and Chen, Jiming},
    title = {A Review on Edge Large Language Models: Design, Execution, and Applications},
    year = {2025},
    publisher = {Association for Computing Machinery},
    address = {New York, NY, USA},
    volume = {57},
    number = {8},
    doi = {10.1145/3719664},
    journal = {ACM Comput. Surv.},
    month = mar,
    articleno = {209},
    numpages = {35}
}

@inproceedings{lau2025adaptive,
    title={Adaptive Batch Size Schedules for Distributed Training of Language Models with Data and Model Parallelism},
    author={Tim Tsz-Kit Lau and Weijian Li and Chenwei Xu and Han Liu and Mladen Kolar},
    booktitle={Proceedings of the Second Conference on Parsimony and Learning (Proceedings Track)},
    year={2025},
}

@inproceedings{lv-etal-2024-coggpt,
    title = "{C}og{GPT}: Unleashing the Power of Cognitive Dynamics on Large Language Models",
    author = "Lv, Yaojia  and
      Pan, Haojie  and
      Wang, Zekun  and
      Liang, Jiafeng  and
      Liu, Yuanxing  and
      Fu, Ruiji  and
      Liu, Ming  and
      Wang, Zhongyuan  and
      Qin, Bing",
    booktitle = "Findings of the Association for Computational Linguistics: EMNLP 2024",
    month = nov,
    year = "2024",
    address = "Miami, Florida, USA",
    publisher = "Association for Computational Linguistics",
    doi = "10.18653/v1/2024.findings-emnlp.352",
    pages = "6074--6091"
}

@inproceedings{echterhoff-etal-2024-cognitive,
    title = "Cognitive Bias in Decision-Making with {LLM}s",
    author = "Echterhoff, Jessica Maria  and
      Liu, Yao  and
      Alessa, Abeer  and
      McAuley, Julian  and
      He, Zexue",
    booktitle = "Findings of the Association for Computational Linguistics: EMNLP 2024",
    month = nov,
    year = "2024",
    address = "Miami, Florida, USA",
    publisher = "Association for Computational Linguistics",
    doi = "10.18653/v1/2024.findings-emnlp.739",
    pages = "12640--12653"
}

@article{10.1145/3757925,
author = {Wang, Xinyuan and Wu, Liang and Hong, Liangjie and Liu, Hao and Fu, Yanjie},
title = {LLM-Enhanced User–Item Interactions: Leveraging Edge Information for Optimized Recommendations},
year = {2025},
publisher = {Association for Computing Machinery},
address = {New York, NY, USA},
volume = {16},
number = {5},
doi = {10.1145/3757925},
journal = {ACM Trans. Intell. Syst. Technol.},
month = sep,
articleno = {117},
numpages = {24}
}

@INPROCEEDINGS{10707432,
  author={Yao, Zhi and Tang, Zhiqing and Lou, Jiong and Shen, Ping and Jia, Weijia},
  booktitle={Proceedings of the 2024 IEEE International Conference on Web Services (ICWS)}, 
  title={VELO: A Vector Database-Assisted Cloud-Edge Collaborative LLM QoS Optimization Framework}, 
  year={2024},
  volume={},
  number={},
  pages={865-876},
}

@ARTICLE{11359542,
  author={Xu, Zhifei and Tang, Zhiqing and Lou, Jiong and Yao, Zhi and Xie, Xuan and Wang, Tian and Wang, Yinglong and Jia, Weijia},
  journal={IEEE Transactions on Mobile Computing}, 
  title={EAT: QoS-Aware Edge-Collaborative AIGC Task Scheduling via Attention-Guided Diffusion Reinforcement Learning}, 
  year={2026},
  volume={},
  number={},
  pages={1-17},
  doi={10.1109/TMC.2026.3656318}}

@misc{zhao2025e2graphragstreamlininggraphbasedrag,
      title={E$^2$GraphRAG: Streamlining Graph-based RAG for High Efficiency and Effectiveness}, 
      author={Yibo Zhao and Jiapeng Zhu and Ye Guo and Kangkang He and Xiang Li},
      year={2025},
      eprint={2505.24226},
      archivePrefix={arXiv},
      primaryClass={cs.AI},
      howpublished={arXiv preprint arXiv:2505.24226}
}

@misc{han2026ragvsgraphragsystematic,
      title={RAG vs. GraphRAG: A Systematic Evaluation and Key Insights}, 
      author={Haoyu Han and Li Ma and Yu Wang and Harry Shomer and Yongjia Lei and Zhisheng Qi and Kai Guo and Zhigang Hua and Bo Long and Hui Liu and Charu C. Aggarwal and Jiliang Tang},
      year={2026},
      eprint={2502.11371},
      archivePrefix={arXiv},
      primaryClass={cs.IR},
      howpublished={arXiv preprint arXiv:2502.11371}
}

@ARTICLE{8657771,
  author={Gao, Guoju and Xiao, Mingjun and Wu, Jie and Huang, He and Wang, Shengqi and Chen, Guoliang},
  journal={IEEE Transactions on Services Computing}, 
  title={Auction-Based VM Allocation for Deadline-Sensitive Tasks in Distributed Edge Cloud}, 
  year={2021},
  volume={14},
  number={6},
  pages={1702-1716},
  doi={10.1109/TSC.2019.2902549}}

\end{document}